\documentclass[letterpaper, 10 pt, conference]{ieeeconf}  

\IEEEoverridecommandlockouts                              

\usepackage{graphicx}
\usepackage{color}
\usepackage{soul}
\usepackage{epsfig} 
\usepackage{mathptmx} 
\usepackage{times} 
\usepackage{amsmath} 
\usepackage{amssymb}  
\usepackage{subfigure}
\usepackage{bm}
\usepackage[export]{adjustbox}
\let\labelindent\relax
\usepackage{enumitem}
\usepackage{multirow}
\usepackage{nicematrix}
\usepackage{float}
\usepackage{xspace}

\newcommand{\cB}{\mathcal{B}}

\newcommand{\cL}{\mathcal{L}}

\newcommand{\cE}{\mathcal{E}}
\newcommand{\cF}{\mathcal{F}}

\newcommand{\cS}{\mathcal{S}}

\newcommand{\bU}{\mathbf{U}}
\newcommand{\bu}{\mathbf{u}}

\newcommand{\bP}{\mathbf{P}}

\newcommand{\bp}{\mathbf{p}}

\newcommand{\bq}{\mathbf{q}}
\newcommand{\bR}{\mathbf{R}}

\newcommand{\bt}{\mathbf{t}}

\newcommand{\be}{\mathbf{e}}

\newcommand{\bK}{\mathbf{K}}

\newcommand{\bOmega}{\bm{\Omega}}
\newcommand{\bpi}{\bm{\pi}}

\newcommand*{\eg}{\emph{e.g.}\@\xspace}
\newcommand*{\ie}{\emph{i.e.}\@\xspace}
\newcommand*{\etal}{\emph{et al.}\@\xspace}
\newcommand*{\cf}{\emph{cf.}\@\xspace}

\definecolor{orbitgreen}{RGB}{158, 224, 136}
\definecolor{approachred}{RGB}{255, 31, 32}

\title{\LARGE \bf
Towards Bridging the Space Domain Gap\\ for Satellite Pose Estimation using Event Sensing}

\author{Mohsi Jawaid$^{\dagger,\ast}$, Ethan Elms$^{\dagger,\ast}$, Yasir Latif$^{\ast}$ and Tat-Jun Chin$^{\ast}$
\thanks{$^{\dagger}$Mohsi Jawaid and Ethan Elms assert equal contributions to the paper.}
\thanks{$^{\ast}$All authors are with the Sentient Satellites Laboratory (SSL) at Australian Institute for Machine Learning (AIML), Adelaide, Australia.}%
}

\begin{document}

\maketitle
\thispagestyle{empty}
\pagestyle{empty}

\begin{abstract}

Deep models trained using synthetic data require domain adaptation to bridge the gap between the simulation and target environments. State-of-the-art domain adaptation methods often demand sufficient amounts of (unlabelled) data from the target domain. However, this need is difficult to fulfil when the target domain is an extreme environment, such as space. In this paper, our target problem is close proximity satellite pose estimation, where it is costly to obtain images of satellites from actual rendezvous missions. We demonstrate that event sensing offers a promising solution to generalise from the simulation to the target domain under stark illumination differences. Our main contribution is an event-based satellite pose estimation technique, trained purely on synthetic event data with basic data augmentation to improve robustness against practical (noisy) event sensors. Underpinning our method is a novel dataset with carefully calibrated ground truth, comprising of real event data obtained by emulating satellite rendezvous scenarios in the lab under drastic lighting conditions. Results on the dataset showed that our event-based satellite pose estimation method, trained only on synthetic data without adaptation, could generalise to the target domain effectively.

\end{abstract}

\section{Introduction}\label{sec:intro}

Object pose estimation is an important capability to enable intelligent robot interactions~\cite{sahin2020review, fan2021deep}. Such a capability is also vital in satellite rendezvous, whereby two or more satellites come into close proximity in orbit to achieve formation flight and/or docking~\cite{zimpfer2005autonomous}, which requires accurate estimation of relative poses between the satellites in close range~\cite{opromolla2017review}.

Robotic vision is a promising approach for object pose estimation. State-of-the-art vision-based pose estimators employ deep learning, whereby a deep neural network (DNN) is trained to predict the pose of the object (or intermediate results such as landmark positions) in an input image~\cite{fan2021deep}. Unsurprisingly, DNN-based visual pose estimation is also being considered actively for satellite rendezvous~\cite{sharma2018pose}.

\begin{figure}[ht]
\subfigure[]{\includegraphics[width=0.24\textwidth,height=0.24\textwidth]{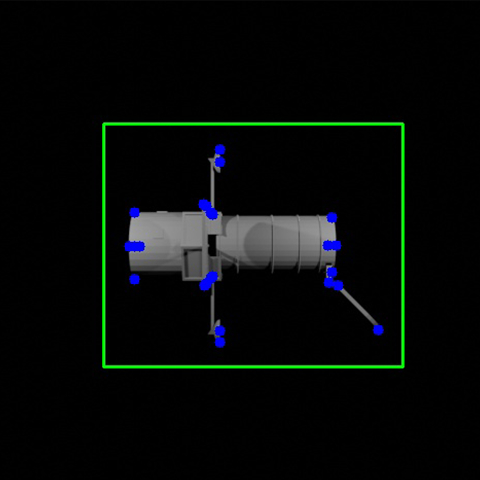}\label{fig:hubble_rgb_synthetic}}
\hfill
\subfigure[]{\includegraphics[width=0.24\textwidth,height=0.24\textwidth]{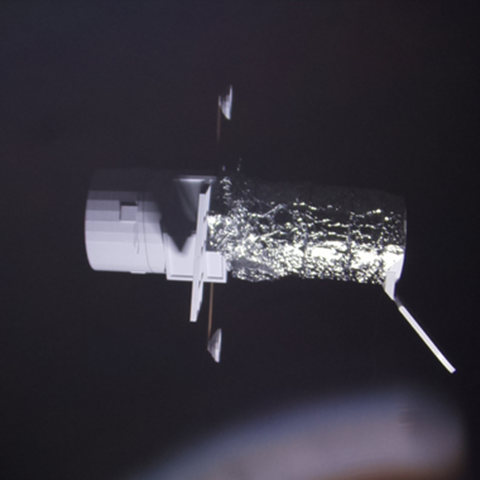}\label{fig:hubble_rgb_real}}

\vspace{-0.7em}

\subfigure[]{\includegraphics[width=0.24\textwidth,height=0.24\textwidth]{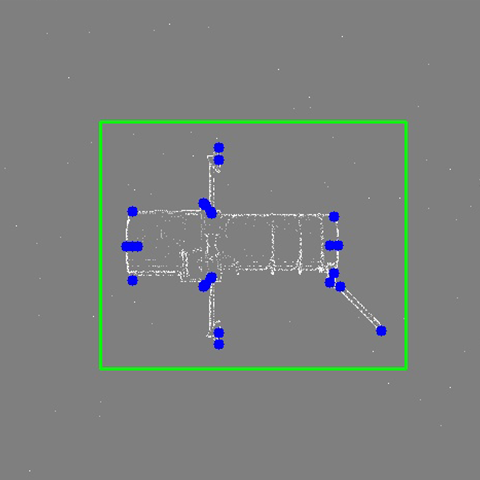}\label{fig:hubble_event_synthetic}}
\hfill
\subfigure[]{\includegraphics[width=0.24\textwidth,height=0.24\textwidth]{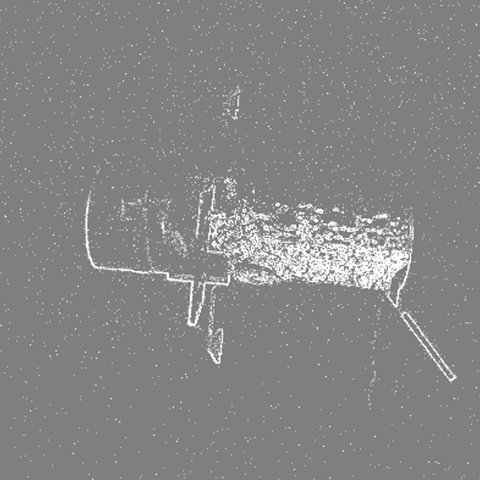}\label{fig:hubble_event_real}}

\vspace{-0.7em}

\caption{(a) Rendered frame of a satellite under normal lighting, with ground truth bounding box and set of $24$ landmarks. (b) Real image of a 3D model of the satellite under extreme lighting; observe the lens flare and uneven contrast. (c) Event frame corresponding to (a), generated using V2E~\cite{hu2021v2e}. (d) Real event frame corresponding to (b). Our premise is that the domain gap between (c) and (d) is lower than that between (a) and (b).}
\vspace{-0.7em}
\end{figure}

However, procuring large amounts of images from actual rendezvous missions to train DNNs is currently a major challenge. Thus, many satellite pose estimation models are trained using computer generated synthetic images~\cite{proencca2020urso, sparkdataset, dung2021spacecraft}; \eg, Fig.~\ref{fig:hubble_rgb_synthetic}. While such an approach allows methods to be rapidly developed and benchmarked~\cite{spec19,spec21}, the resulting models cannot transfer to the target domain. To alleviate this problem, visual domain adaptation (VDA)~\cite{WANG2018135,peng2018visda} is vital to bridge the simulation-to-real (Sim2Real) gap~\cite{park2021speedplus}.

Major techniques for VDA nevertheless need target domain images, in addition to the source domain (synthetic) images~\cite{WANG2018135,peng2018visda}. In the case of unsupervised VDA, \emph{unlabelled} target domain images are required to transductively learn DNNs that can generalise to the target domain. However, as alluded to above, acquiring images from real rendezvous missions is difficult. Therefore, recent studies on VDA for satellite pose estimation~\cite{park2021speedplus} relied on in-laboratory physical emulation to produce target domain images, which arguably still suffers from the Sim2Real gap.

Being an extreme environment, the lighting conditions in space present significant challenges to robotic perception algorithms, \eg, high contrast, camera over-exposure, shadowing and stray light~\cite{park2021speedplus}; see Fig.~\ref{fig:hubble_rgb_real}. Moreover, accurately emulating space lighting conditions in the lab is non-trivial, which compounds the difficulty of Sim2Real. While VDA will remain an important tool, the challenges from space motivate considering other approaches.

In this paper, we explore satellite pose estimation that is inherently robust to extreme lighting. We leverage the high dynamic range and asynchronous change detection of event sensors to perceive satellites in a way that is insensitive to drastic illuminations. We then train a satellite pose estimator from \emph{synthetic} event data, processed from 3D rendering of the satellite under mild lighting conditions, with data augmentations to deal with noise expected in real devices. We show that the synthetically trained pose estimator can transfer well to the real domain \emph{without} VDA, since event data is less affected by illumination changes, \cf~Figs.~\ref{fig:hubble_event_synthetic} and~\ref{fig:hubble_event_real}. This helps to bridge the Sim2Real gap.

While our satellite pose estimation pipeline is not entirely novel, we are the first to investigate it on event data. Another major contribution is a calibrated event dataset~\cite{ourdataset} acquired by robotic emulation of satellite rendezvous using a state-of-the-art event camera, under carefully controlled extreme lighting. While our event dataset still suffers from the space domain gap, it nevertheless provides a strong basis to demonstrate Sim2Real transfer by our method across real and simulated domains with very different lighting conditions.

\section{RELATED WORK}

\subsection{Satellite pose estimation}

Estimating the 6DoF pose (position and orientation) of a target object relative to an observing platform is fundamental to robotics~\cite{fan2021deep, sahin2020review}. The specific version of interest in this paper is satellite pose estimation during in-space rendezvous and docking, which underpin applications such as on-orbit refuelling, on-orbit maintenance and debris removal~\cite{shoemaker2020osam}.

When the target satellite is uncooperative (\eg, a defunct satellite or space debris), pose estimation must be accomplished by the chaser satellite independently. Visual pose estimation using a single camera is attractive due to its simpler design. The importance of the topic is underlined by several datasets \cite{speed2020,park2021speedplus,proencca2020urso} and challenges \cite{spec19,spec21,sparkchallenge21}. Following the success of DNN-based object pose estimation in computer vision \cite{fan2021deep}, state-of-the-art monocular satellite pose estimation methods are also based on DNNs \cite{sharma2019spn,bospec2019}.

\subsection{Visual domain adaptation}

VDA is crucial to enable models trained on image data from a source domain to work in a target domain~\cite{WANG2018135,peng2018visda}. The difference between the two domains is called the \emph{domain  gap}~\cite{csurka2017domain}. Of particular interest is when the source domain is a simulation where it is possible to generate vast quantities of labelled synthetic images. The Sim2Real gap~\cite{hofer2021sim2real,peng2018syn2real} must be overcome using VDA to enable models trained on synthetic data to operate well in the real domain.

Addressing the Sim2Real gap is also crucial in satellite pose estimation, since many methods have been developed based on synthetic data~\cite{proencca2020urso, sparkdataset, dung2021spacecraft,sharma2019spn,bospec2019}. In addition, the real space environment is affected by extreme lighting conditions~\cite{https://doi.org/10.48550/arxiv.2203.04275}, which is difficult to simulate virtually. A major recent effort to address Sim2Real for satellite pose estimation is by Park \etal~\cite{park2021speedplus}, who developed a dataset (SPEED+) that contained synthetic training images with ground-truth pose labels as well as unlabelled real testing images captured in a lab with challenging lighting conditions. SPEED+ formed the basis for the recent Kelvins Satellite Pose Estimation Challenge 2021~\cite{spec21}. However, as alluded to in Sec.~\ref{sec:intro}, the dataset still suffers from domain gap since the real images are not from the actual space environment.

\subsection{Event-based vision}

In the context of robotic vision, event sensors offer several advantages over RGB sensors, such as higher dynamic range and asynchronous operation. Many recent works have exploited event sensors for robotic vision tasks such as object recognition, classification, semantic segmentation, VO and SLAM~\cite{perot2020learning, martin2021evaluation, mahlknecht2022exploring, jia2022event, xiao2022research}.

Relatively less attention has been paid to pose estimation from event data. Reverter Valeiras \etal~\cite{reverter2016neuromorphicpose} proposed an asynchronous technique to update a given pose of an object from the event stream. Nguyen~\etal~\cite{nguyen2017real} investigated camera relocalisation from event data~\cite{nguyen2017real}. Chen \etal~\cite{chen2022efficient} explored articulated human pose estimation. Our work differs from the above since we estimate the object pose without prior knowledge of the pose and under strong illumination effects.

In the space domain, applications of event sensors are just emerging. Early works include star tracking~\cite{chin2019star} and odometry for planetary robots~\cite{mahlknecht2022exploring} using event sensing.

\subsection{Event datasets}

In line with the rapidly growing interest in event sensing for robotic vision, event datasets are increasingly being produced. However, many of the datasets support the tasks of object detection, object classification, VO and SLAM~\cite{mitrokhin2018event, DBLP:journals/corr/abs-1803-07913, Bryner19icra, DBLP:journals/corr/MuegglerRGDS16,gehrig2020video, gehrig2019video}. The event dataset~\cite{ourdataset} we contribute here is one of the first on satellite pose estimation.

\begin{figure*}
\includegraphics[width=2\columnwidth,height=0.45\columnwidth]{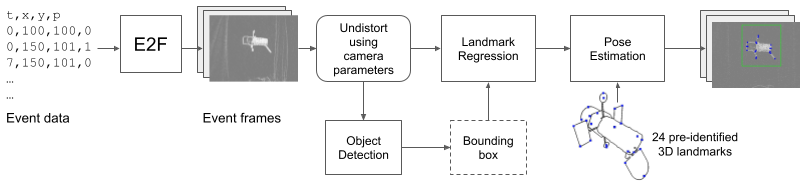}
\vspace{-1em}
\caption{Proposed pipeline for satellite pose estimation from event frames.}
\label{fig:pose_estimation_pipeline}
\end{figure*}

\section{Satellite pose with event sensing}\label{sec:method}

This section describes our method of conducting satellite pose estimation using event data. Similar to the setting in~\cite{speed2020,park2021speedplus,proencca2020urso,hu2021wide,pasqualetto2020cnn,sparkdataset}, we assume that the target satellite is uncooperative, but we have knowledge of the structure of the satellite (\eg, via CAD model or SfM reconstruction).

Before proceeding, we emphasise that our pipeline will be trained on only synthetic data, and tested on real data without VDA. Details on our dataset and experiments will be provided respectively in Secs.~\ref{sec:dataset} and~\ref{sec:results}.

\subsection{From event streams to event frames}\label{sec:batching}

An event stream $\cE$ has the form $\cE = \{ \be_1, \be_2, \dots \}$, where each event $\be_i = \{x_i,y_i,p_i,t_i\}$ is a tuple with image coordinates $(x_i,y_i)$, polarity $p_i \in \{-1,1\}$ and timestamp $t_i$. To leverage existing DNN methods for satellite pose estimation~\cite{sharma2019spn,bospec2019}, we convert $\cE$ into event frames $\cF = \{ I_1, I_2, \dots \}$ via event-to-frame (E2F) conversion, where each $I_j \in [0,\Gamma]^{M \times N}$ is an image obtained from an event batch $\cB_j \subset \cE$ within a specific time window $W_j$, $\Gamma$ is the highest possible intensity, and $M \times N$ is the spatial resolution of the event sensor.

Our specific E2F method creates for each input batch $\cB_j$ a 2D histogram with $M \times N$ cells of the image coordinates $(x_i,y_i)$ of the events in $\cB_j$. The values of the histogram are then normalised ($\Gamma = 1$) and exported as an intensity image; see Figs.~\ref{fig:hubble_event_synthetic} and~\ref{fig:hubble_event_real}. An important parameter in E2F is the duration $\tau$ of each $\cB_j$; this will be discussed in Sec.~\ref{sec:hyperparam}.

\subsection{Pose estimation from event frames}

Let $\cS = \{ \bP_k \}_{k=1}^{K}$ be a 3D point cloud that defines the structure of the target satellite in a canonical reference frame. Given an input event frame $I_j$ that observed the satellite in time window $W_j$, the projection of $\cS$ onto $I_j$ is given by
\begin{align*}
    \tilde{\bp}_k = \bK[\mathbf{R}_j~\bt_j]\tilde{\bP}_k,~~~k = 1,\dots,K, 
\end{align*}
where $\bK$ contains the camera intrinsics, and $\tilde{\bp}_k$ is $\bp_k$ in homogeneous coordinates (similarly for $\tilde{\bP}_k$). The rigid transformation $(\mathbf{R}_j,\bt_j)$ defines the pose of $\cS$ in event frame $I_j$, or, alternatively, during the time window $W_j$ that generated the event batch $\cB_j$.

\textbf{Why use event frames?} Strictly speaking, the notion of a ``static'' pose $(\bR_j,\bt_j)$ within a time duration $W_j$ is imprecise due to relative motion of the camera and object in $W_j$; indeed, the event camera needs to move to generate events. However, $W_j$ is usually small (\eg, $<1$ s) relative to the motion speed, hence $(\bR_j,\bt_j)$ is sufficient to describe the pose. While a fully asynchronous treatment of pose estimation is ideal, our event frame approach is sufficient for our primary aim: \emph{demonstrate Sim2Real transfer for satellite pose estimation}.

\subsection{Proposed method}

Fig.~\ref{fig:pose_estimation_pipeline} shows our pipeline, which was adapted from our winning solution~\cite{spec21results} to the Kelvins Satellite Pose Estimation Competition 2021~\cite{spec21} (\texttt{Lightbox} category). The main differences between Fig.~\ref{fig:pose_estimation_pipeline} and the previous method are
\begin{itemize}[leftmargin=1em]
\item Using event frames instead of RGB frames; and
\item Removing all VDA steps (\eg, adversarial training) that were introduced to bridge the Sim2Real gap in~\cite{spec21}.
\end{itemize}
Details of our method are given below.

\subsubsection{Landmark selection}

A small subset $\cL = \{ \bU_z \}^{Z}_{z = 1}$ of $\cS$ that represent salient points on the surface of the object was first selected; see Fig.~\ref{fig:pose_estimation_pipeline}. The 2D positions of the landmarks $\cL$ in an input event frame will be the target of prediction by the landmark regressor (see below).

\subsubsection{Object detection}

We used Detectron2~\cite{wu2019detectron2} with Faster-RCNN~\cite{ren2015fasterrcnn} + FPN~\cite{Lin_2017_CVPR} backbone for the object detection task, which was sufficiently accurate for our pipeline (single instance detection on a sparse background).

\subsubsection{Landmark regression}

We use HRNet \cite{sun2019hrnet} with $512\times 512$ images and $128\times 128$ heatmaps to predict the 2D positions $\{ \bu_z \}^{Z}_{z = 1}$ of the landmarks $\cL$ within the detected bounding box of the target satellite in the input event frame.

\subsubsection{Pose estimation}

The 2D-3D correspondences $\{ (\bu_z, \bU_{z} ) \}^{Z}_{z = 1}$ form the input to a perspective-n-points (PnP) solver to estimate the object pose $(\bR,\bt)$ corresponding to the input event frame. To improve robustness and accuracy, we filter the 2D-3D correspondences based on the per-point confidence score output from HRNet such that at least 15 correspondences with a score of more than a 0.95 threshold are kept. If there are not enough such points, we reduce the threshold by 20\% until at least 15 points are obtained. 

\subsubsection{Training data}

Synthetic event frames with ground truth bounding boxes and 2D landmarks (see Sec.~\ref{sec:dataset}) were employed to train the object detector and landmark regressor.

\subsubsection{Training}\label{sec:training}

Training was performed on a single NVIDIA RTX 3090 24GB graphics card. For object detection we used the SGD optimizer which runs for 10000 epochs with a batch size of 10 images. An initial learning rate of 0.0001 is used which is reduced by a factor of 0.1 after 8000 steps. HRNet was trained for 40 epochs with a batch size of 24. The \texttt{adam}~\cite{kingma2014adam} optimizer was used with the initial learning rate set to 0.001 which is reduced by a factor of 0.1 after the 25th and 35th epoch.

\subsubsection{Data augmentation}

\begin{figure}[ht]\centering
\subfigure{\includegraphics[width=0.15\textwidth,height=0.18\textwidth]{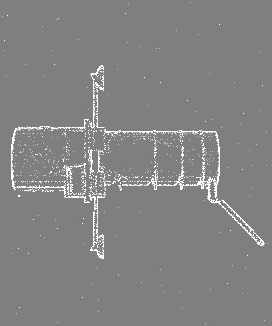}\label{fig:augmentation_normal}}
\subfigure{\includegraphics[width=0.15\textwidth,height=0.18\textwidth]{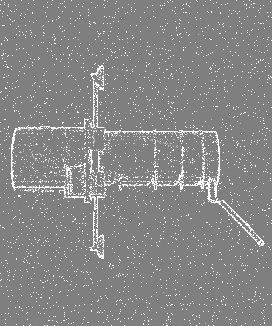}\label{fig:augmentation_noise}}
\subfigure{\includegraphics[width=0.15\textwidth,height=0.18\textwidth]{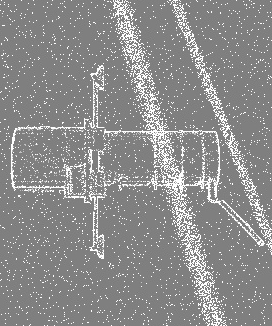}\label{fig:augmentation_noise_and_lines}}
\caption{Our augmentations for event frames. (Left) Event frame generated using V2E. (Middle) With \texttt{RandomEventNoise} augmentation. (Right) With \texttt{RandomEventNoise} and \texttt{RandomEventLines} augmentations.}
\label{fig:augmentations}
\end{figure}

We subjected the training data to random rotation and random translation data augmentations. Furthermore, we also used our custom augmentations for event-frames called \texttt{RandomEventNoise} and \texttt{RandomEventLines} (see Fig.~\ref{fig:augmentations}) to mitigate the effects of background artefacts in our real data capture environment (Sec.~\ref{sec:dataset}). All four augmentations were used for both the object detection and landmark regression phases. 

\subsubsection{Testing data}
\label{sec:testing}

Real event data were procured to evaluate the pipeline (details in Sec.~\ref{sec:dataset}).

\section{Event dataset for satellite pose estimation}\label{sec:dataset}

To demonstrate Sim2Real transfer for satellite pose estimation using event sensing, we produced synthetic and  real event datasets, which will be publicly released~\cite{ourdataset}.

\subsection{Synthetic event data}\label{sec:synthetic generation}

\subsubsection{Simulation environment}
\begin{figure}[ht]\centering
\vspace{-0.5em}
\subfigure{\includegraphics[width=0.22\textwidth,height=0.18\textwidth]{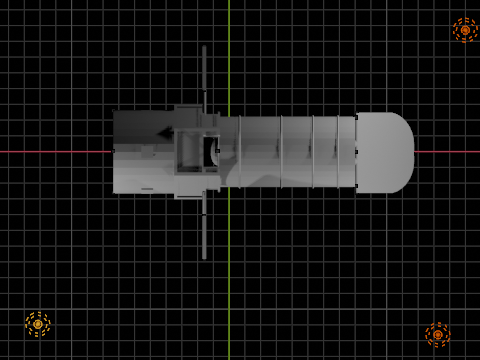}\label{fig:synthetic_lighting_top_rendered}}
\hspace{0.5em}
\subfigure{\includegraphics[width=0.22\textwidth,height=0.18\textwidth]{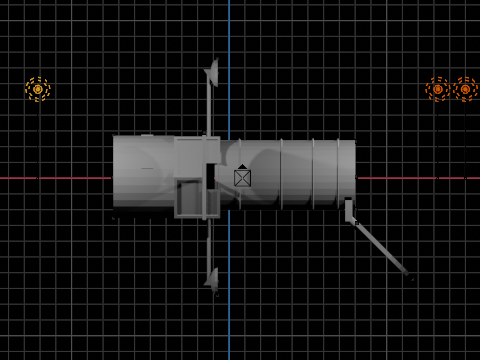}\label{fig:synthetic_lighting_side_rendered}}\\
\vspace{-0.5em}
\caption{Blender viewport with the HST 3D model and three point light sources (orange) for illumination. (Left) Top view and (Right) side view.}
\label{fig:synthetic_lighting}
\end{figure}

A textureless 3D model of the Hubble Space Telescope (HST)~\cite{hubble} was rendered in Blender~\cite{blender}. The HST was specifically chosen due to its complex structure of protruding, non-uniform elements, which creates self occlusions and shadows. The virtual camera follows a smooth trajectory while observing the HST from varying poses. Only \emph{one} lighting condition was used, where we placed 3 point light sources around the satellite at the same height; see Fig.~\ref{fig:synthetic_lighting}. The setup was chosen to evenly illuminate the object with mild shadows; see Fig.~\ref{fig:hubble_rgb_synthetic}.

Following Sec.~\ref{sec:method}, $24$ salient points distributed across the 3D surface of the HST were manually chosen as the landmark set $\cL$; see Fig.~\ref{fig:pose_estimation_pipeline}. Based on this setup, an RGB video with 10k frames of $640\times 480$ pixels was rendered (the resolution matches the real event camera Inivation DVXplorer employed in the real data collection; details in Sec.~\ref{sec:event-camera}). A high frame rate of $100$ FPS was used to facilitate synthetic event generation (details in Sec.~\ref{sec:syn_frames_to_events}).

\subsubsection{Ground truth}

The intrinsics and pose of the virtual camera in each RGB frame in the video were exported from Blender. The image positions of the landmarks $\cL$ were obtained by projecting their 3D coordinates onto the RGB frames. From the landmark positions, a bounding box with sufficient clearance (10\% larger than the tightest bounding box on the landmarks) was defined; see Fig.~\ref{fig:hubble_rgb_synthetic}.

\subsubsection{Synthetic frames to events}\label{sec:syn_frames_to_events}

The rendered RGB video frames were subjected to V2E~\cite{hu2021v2e} which employed a method similar to Katz~\etal~\cite{v2e2012katz} to convert linear (0-255) intensity RGB frames to log intensity to simulate event sensors. We used a minimum timestamp resolution of $0.01$ s to match the frame rate of the rendering. The high frame-rate obviated the need for additional frame interpolation in V2E.

From the generated event data, we produced event frames following the E2F procedure in Sec.~\ref{sec:batching}. To increase robustness of the trained model towards different batching durations, we used multiple batching durations $\tau$ ($0.2$, $0.1$, $0.05$ and $0.01$ seconds). For each event frame, we assigned the ground truth pose, landmark positions and bounding box of the RGB frame with the closest timestamp. Fig.~\ref{fig:hubble_event_synthetic} shows an event frame processed from our synthetic event data.

\subsection{Real event data}\label{sec:realdata}

\begin{figure}[ht]\centering
\subfigure{\includegraphics[height=0.24\textwidth]{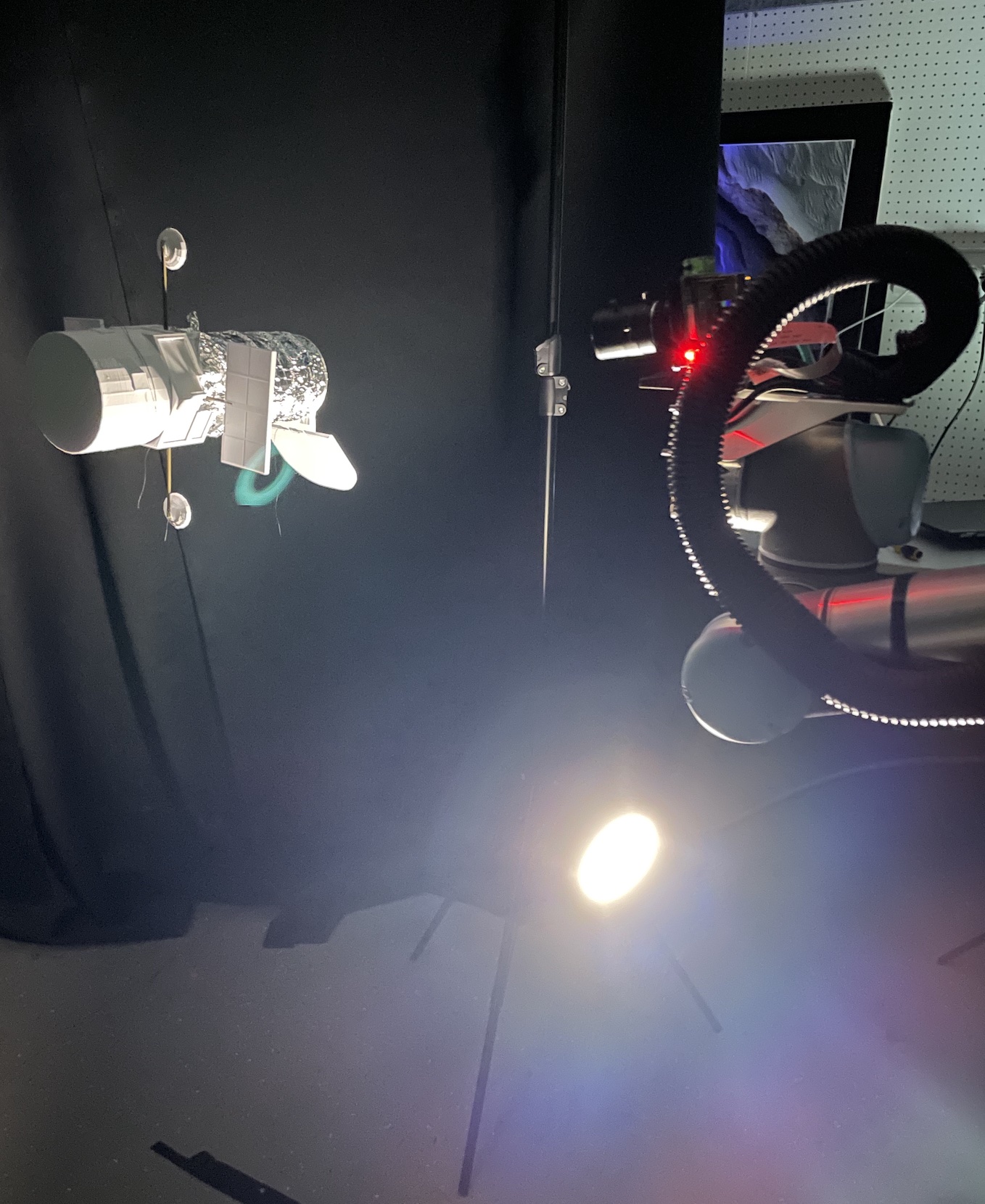}\label{fig:hubble_setup_tall}}
\hspace{1em}
\subfigure{\includegraphics[ height=0.24\textwidth]{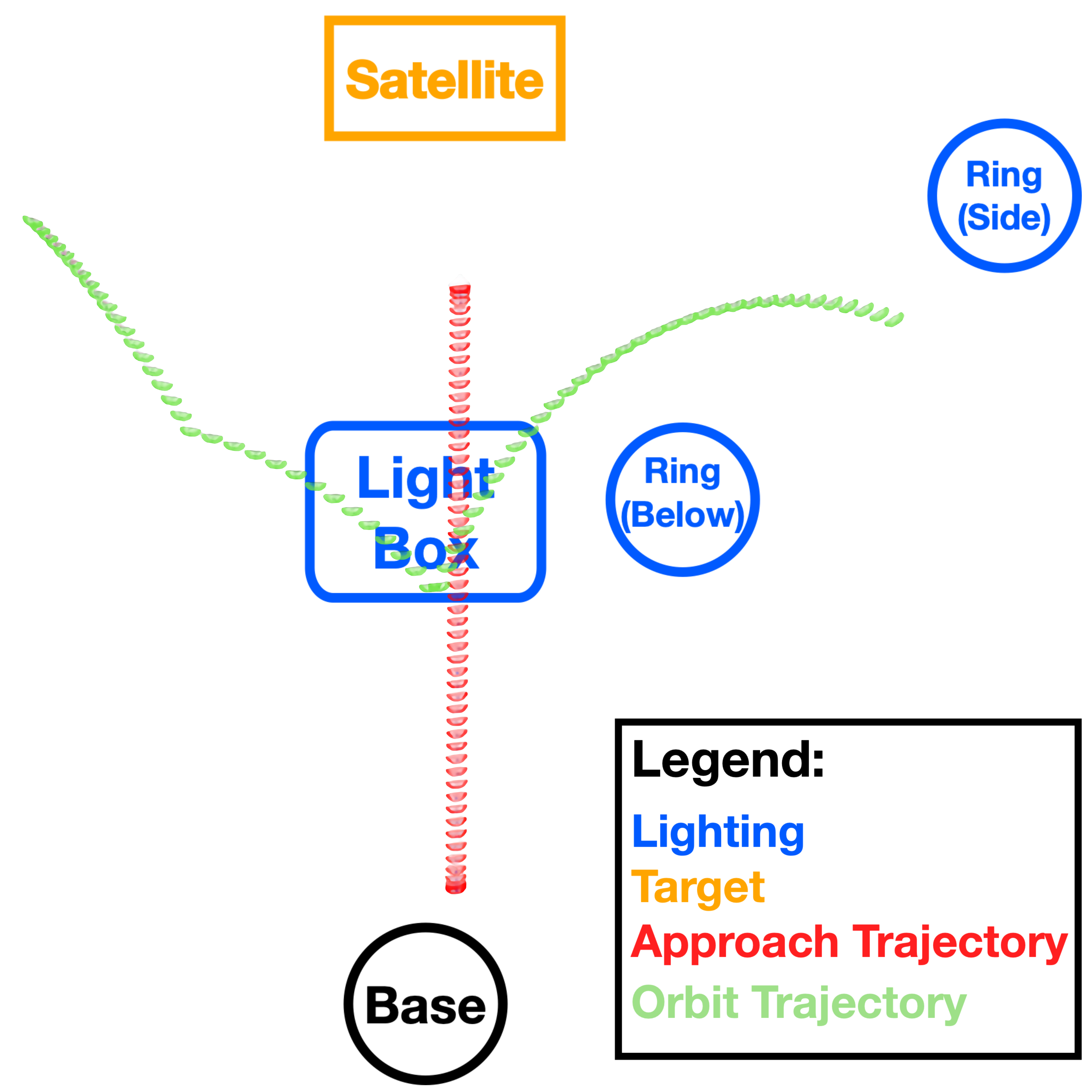}\label{fig:trajectory_poses}}\\
\vspace{-0.5em}
\caption{(Left) Our setup for capturing real event data. Shown here is the \texttt{ringbelow} lighting configuration. (Right) Distribution of ground truth event camera poses for the \texttt{approach} (\textcolor{approachred}{red}) and \texttt{orbit} (\textcolor{orbitgreen}{green}) trajectories with approximate light source positions and directions (not to scale).}
\end{figure}

A real event dataset was captured using an event camera mounted on a UR5e robot arm to observe a printed 3D model of the HST~\cite{hubble}; see Fig.~\ref{fig:hubble_setup_tall}. Details are provided below.

\subsubsection{Printed 3D model}

To mimic uneven reflectance and texture on the real HST, we partially wrapped our model with aluminium foil to emulate a thermal blanket; see Fig.~\ref{fig:hubble_rgb_real}.

\subsubsection{Event camera}\label{sec:event-camera}

We employed an Inivation DVXplorer event camera, with a Arducam 4-12mm Varifocal C-Mount lens which has an adjustable focal length via manual focus and aperture rings. To focus the event camera, it was placed at the initial pose of the approach trajectory, and a rotating Siemens Star was placed near the satellite. The focus on the lens was then adjusted manually until the pattern was sharp.

\subsubsection{Event camera calibration}
\label{sec:event-camera-calib}

We adapted the intrinsic and extrinsic (eye-in-hand) calibration techniques for RGB cameras~\cite{tsai_lenz_1989,zhang_2000}, as implemented in OpenCV~\cite{opencv_library}. First, we executed a semi-spherical trajectory on the UR5e arm, whilst carrying the event camera to observe a checkerboard on an LED screen. The observed events and the ground truth robot arm poses (polled at $10$ Hz using the manufacturer's API) served as inputs to the calibration. We accumulated the events into event frames using the Inivation DV SDK's accumulation frame module, then ran chessboard corner detection. The results were passed to \texttt{calibrateCamera}, which yielded the camera intrinsics. Using the intrinsics, the chessboard to camera transformation was estimated using \texttt{solvePnP}, which was then used in \texttt{calibrateHandEye} as the \texttt{target2cam} input transformation. See~\cite{ourdataset} for detailed information of the calibration. 

\subsubsection{Rendezvous trajectories}

Two trajectories named \texttt{approach} and \texttt{orbit} were executed; see Fig~\ref{fig:trajectory_poses}. The first one mimics a docking procedure with the satellite, where the camera makes a direct, linear approach towards the satellite; the second mimics an orbit around the satellite. Two different speeds (\texttt{slow} and \texttt{fast}) were also activated for each trajectory. Specifically, \texttt{approach-slow} at $0.0332$ m/s, \texttt{orbit-slow} at $0.0142$ m/s, \texttt{approach-fast} at $0.2186$ m/s, and \texttt{orbit-fast} at $0.3007$ m/s.

\subsubsection{Lighting conditions}

Four lighting configurations were tested: \texttt{ambient} was the standard indoor lights of our lab, which illuminate the entire scene evenly. A $9300$ Lumens light, combined with a custom light diffuser, was used to create the \texttt{lightbox} lighting.
The \texttt{centre} lighting used a portion of the lab lighting array, with the intention to provide a dim illumination of the entire scene.
Lastly, two small ring-shaped lights were employed for the \texttt{ringside} and \texttt{ringbelow} lightings, where the lights were situated to the right side and below the approach trajectory, respectively.

\begin{figure}[ht]\centering
\includegraphics[height=0.24\textwidth]{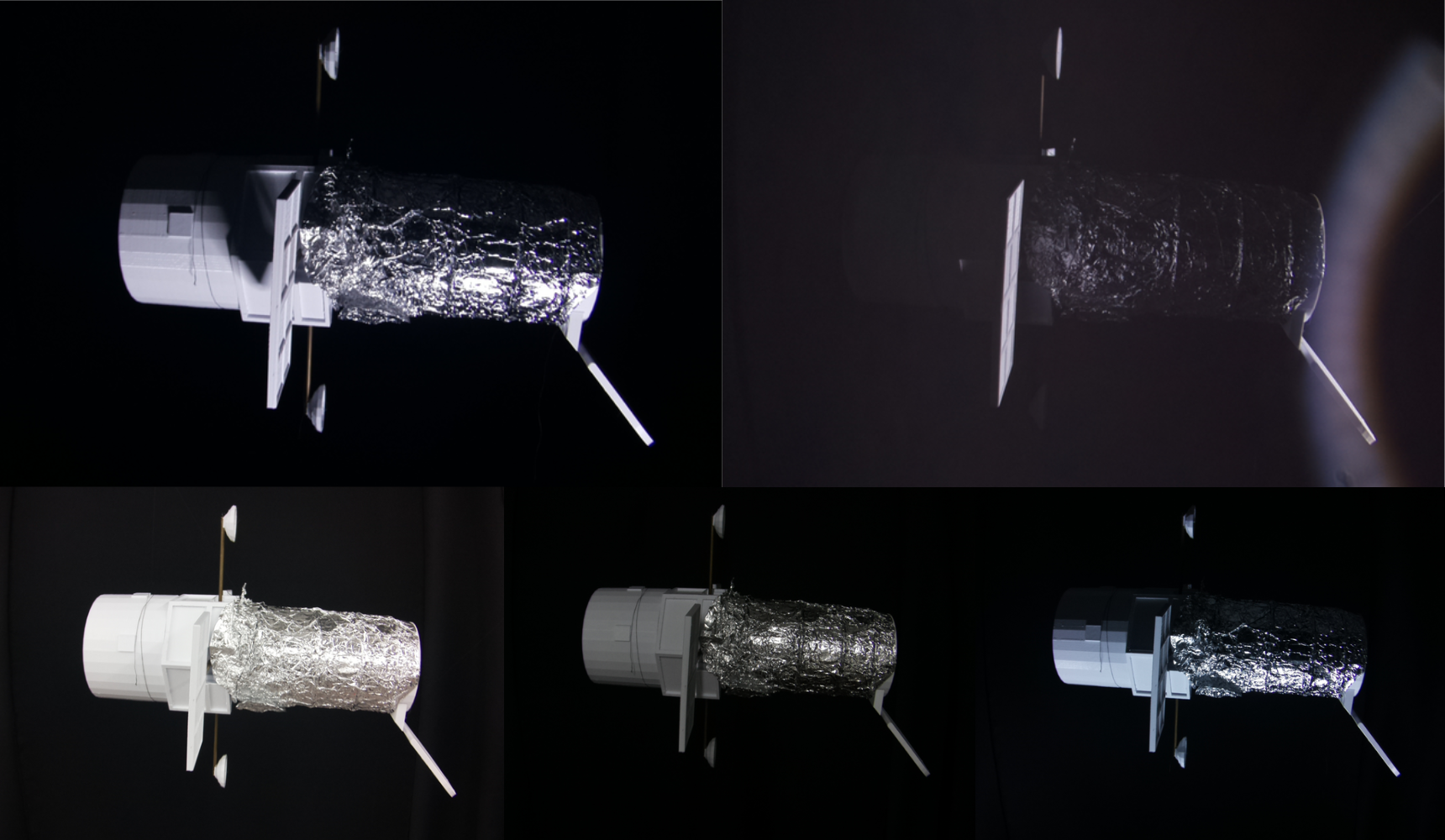}
\caption{Lighting effects on satellite model (from top to bottom, left to right: \texttt{ringbelow}, \texttt{ringside}, \texttt{ambient}, \texttt{centre}, \texttt{lightbox}).}
\label{fig:rgb_lighting_conditions}
\end{figure}

The location of the light sources relative to the camera motion and target object are shown in Fig.~\ref{fig:trajectory_poses}, while Fig.~\ref{fig:rgb_lighting_conditions} illustrates the lighting effects via RGB still images.

\subsection{Statistics of the real event dataset}

\begin{figure}[ht]
\includegraphics[width=0.49\textwidth]{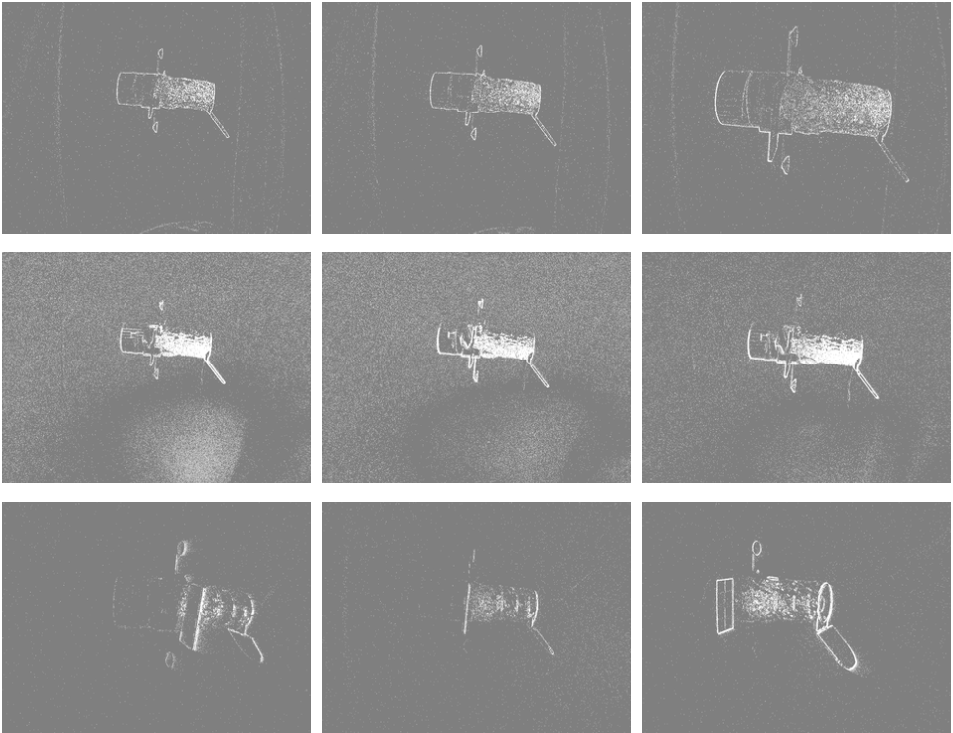}
\caption{Sample real event frames from our dataset. \texttt{approach-slow-ambient} (top), \texttt{approach-fast-ringbelow} (middle) and \texttt{orbit-fast-ringside} (bottom). Note that the vast lighting differences did not produce noticeable effects in the data, \cf~Fig.~\ref{fig:rgb_lighting_conditions}.}
\label{fig:real_event_frames}

\end{figure}

All combinations of trajectory type, speed and lighting configuration were enumerated for capture. The capture output for each sequence is an event stream $\cE$ and ground truth camera poses $\{ (\bOmega_\ell,\bpi_\ell) \}_{\ell=1}^{L}$ at $10$ Hz. Fig.~\ref{fig:real_event_frames} illustrates real event frames from our dataset, while Table~\ref{tab:dataset_statistics} provides an overview of the collected real event data.


Our dataset for \textbf{Spacecraft posE Estimation with NeuromorphIC vision (SEENIC)}, will be released publicly~\cite{ourdataset}.

\begin{table}[H]
\centering
\begin{NiceTabular}{llccc}[hvlines,rules/color=[gray]{0.3}]
Trajectory & Scene & Events & Time (s) & Cam.~poses \\
\Block{10-1}{approach} & fast-ambient & 1.7 M & 2.40 & 23 \\
& fast-centre & 1.5 M & 2.40 & 23 \\
& fast-lightbox & 7.5 M & 2.41 & 23 \\
& fast-ringbelow & 4.1 M & 2.40 & 23 \\
& fast-ringside & 1.6 M & 2.40 & 23 \\
& slow-ambient & 6.3 M & 14.73 & 146 \\
& slow-centre & 5.8 M & 14.73 & 146 \\
& slow-lightbox & 43.3 M & 14.73 & 146 \\
& slow-ringbelow & 10.3 M & 14.72 & 146 \\
& slow-ringside & 3.9 M & 14.73 & 146 \\
\Block{10-1}{orbit} & fast-ambient & 6.5 M & 4.31 & 42 \\
& fast-centre & 6.5 M & 4.31 & 42 \\
& fast-lightbox & 17.1 M & 4.31 & 42 \\
& fast-ringbelow & 11.8 M & 4.31 & 42 \\
& fast-ringside & 4.7 M & 4.31 & 42 \\
& slow-ambient & 39.2 M & 87.48 & 872 \\
& slow-centre & 31.9 M & 87.46 & 872 \\
& slow-lightbox & 276.7 M & 88.74 & 872 \\
& slow-ringbelow & 52.7 M & 87.48 & 872 \\
& slow-ringside & 16.3 M & 87.48 & 872 \\
\end{NiceTabular}
\caption{Basic statistics of our real event dataset (M = Million).}
\label{tab:dataset_statistics}
\end{table}

\section{Results}\label{sec:results}
\subsection{Hyperparameters}\label{sec:hyperparam}

The settings of the important hyperparameters of our method were as follows:
\begin{itemize}[leftmargin=1em]
    \item Batch duration $\tau = 0.05$ s for the \texttt{approach} scenarios, $0.2$ s for the \texttt{orbit-slow} scenarios and $0.01$ s for the \texttt{orbit-fast} scenarios. These values were selected to optimise the signal-to-noise ratio in the event frames.
    \item $Z = 24$ points on the surface of HST that cover unique structures such as the hatch, solar panels and dishes were selected as the landmark set $\cL$; see Fig.~\ref{fig:pose_estimation_pipeline}.
\end{itemize}

\subsection{Qualitative results}

\begin{figure}[ht]
\includegraphics[width=0.99\columnwidth,height=0.50\columnwidth]{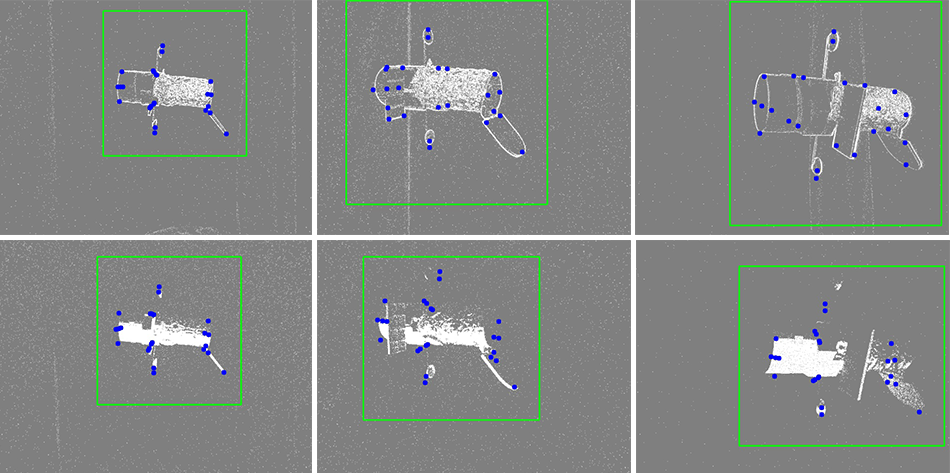}
\caption{Sample of real event frames with the predicted object detection bounding box in green and 3D landmarks reprojected using the estimated pose in blue. In each row we show a frame from the \texttt{approach-slow}, \texttt{orbit-slow} and \texttt{orbit-fast} scenes respectively in different lighting. \texttt{ambient} (top) \texttt{lightbox} (bottom).}
\label{fig:real_event_results}
\end{figure}

Fig.~\ref{fig:real_event_results} shows sample outputs of our method on real data (Sec.~\ref{sec:realdata}), where the method was trained on only synthetic data (Sec.~\ref{sec:synthetic generation}). Generally the position estimates are better than the orientation estimates; the error in the latter likely due to slight misalignments with the solar panels.

\subsection{Quantitative results}

Each real event stream $\cE$ was converted into event frames $\{ I_1, I_2, \dots \}$ following Sec.~\ref{sec:batching}. Then, each ground truth camera pose $(\bOmega_\ell,\bpi_\ell)$ for $\cE$ was paired with the event frame $I_j$ that was closest in time. The trained model was executed on $I_j$ to estimate the object pose $(\mathbf{R}_j,\bt_j)$. The process created $L$ pairs of poses $\{(\bOmega_\ell,\bpi_\ell)\}^{L}_{\ell=1}$ and $\{ (\mathbf{R}_\ell,\bt_\ell) \}_{\ell=1}^{L}$.

To measure the accuracy of object pose estimation, we computed the relative trans.~between successive poses
\begin{align*}
(\bR^{rel}_\ell,\bt^{rel}_\ell) &= \texttt{rel(}(\bR_\ell,\bt_\ell), (\bR_{\ell+1},\bt_{\ell+1})\texttt{)},\\
(\bOmega^{rel}_\ell,\bpi^{rel}_\ell) &= \texttt{rel(}(\bOmega_\ell,\bpi_\ell), (\bOmega_{\ell+1},\bpi_{\ell+1})\texttt{)}.
\end{align*}
Following~\cite{park2021speedplus}, we compute the error of the object pose estimation via the error in the relative transformation, \ie,
\begin{align*}
\phi_\ell = \left\| \bt^{rel}_\ell - \bpi^{rel}_\ell \right\|_2,~~~~\psi_\ell =  2\arccos \left( | <\bq^{rel}_\ell, \hat{\bq}^{rel}_{
\ell}> | \right),
\end{align*}
which resp.~measure translation error (in meters) and rotation error (in degrees), and $\bm{q}^{rel}_\ell$ and $\hat{\bm{q}}^{rel}_{\ell}$ are the quaternion form of $\bR^{rel}_\ell$ and $\bOmega^{rel}_\ell$. The overall error for $\cE$ is
\begin{align*}
\Phi = \sqrt{\frac{1}{L-1}\sum_{\ell=1}^{L-1} \phi_\ell^2},~~~~\Psi = \frac{1}{L-1}\sum_{\ell=1}^{L-1} \psi_\ell,
\end{align*}
which are also in meters and degrees, respectively.

\begin{figure}[ht]\centering
    \subfigure{\includegraphics[width=0.5\textwidth]{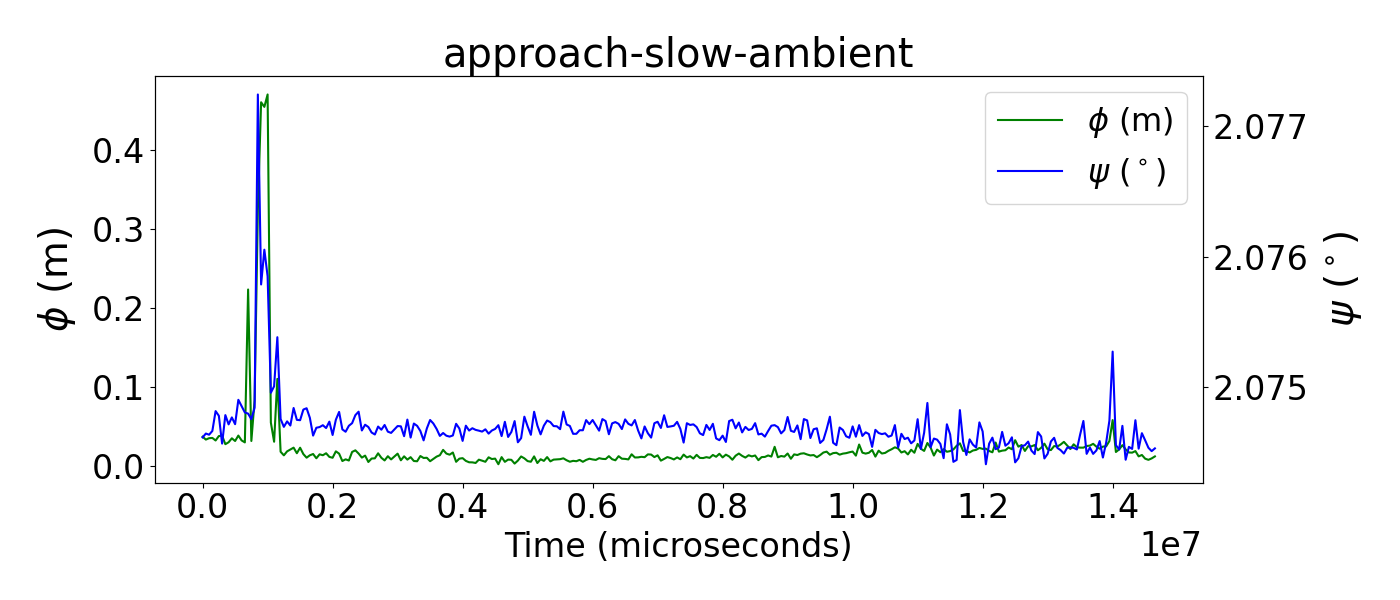}}
    \subfigure{\includegraphics[width=0.5\textwidth]{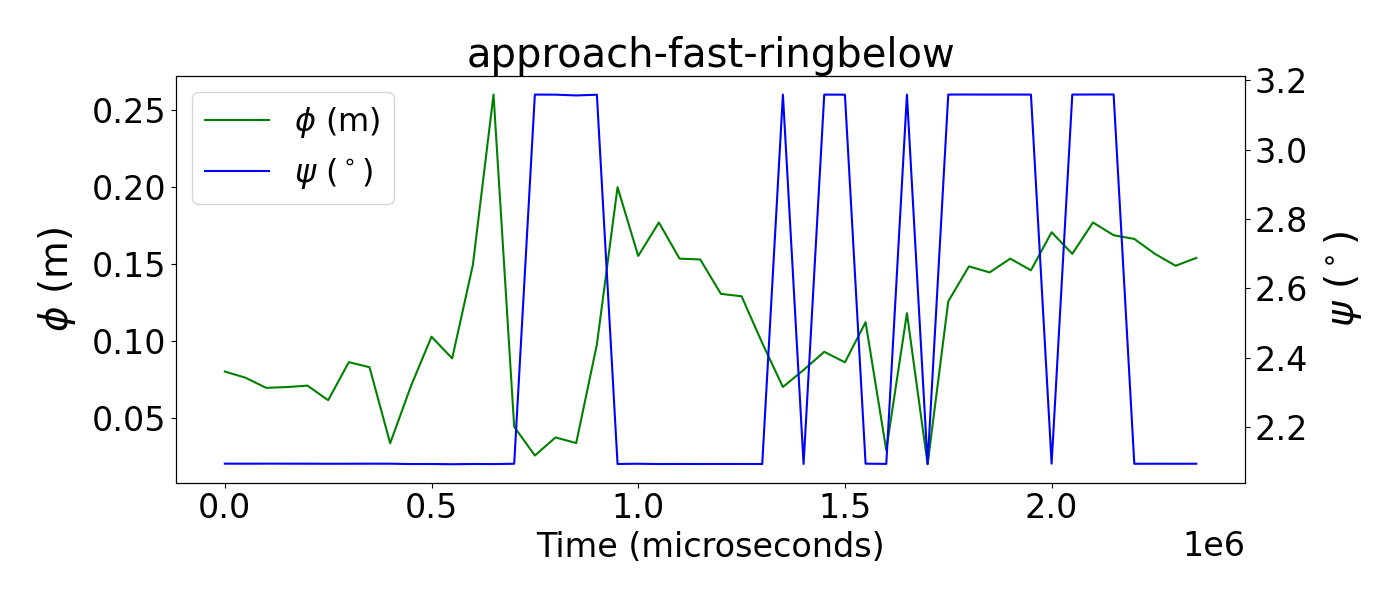}}
    \subfigure{\includegraphics[width=0.5\textwidth]{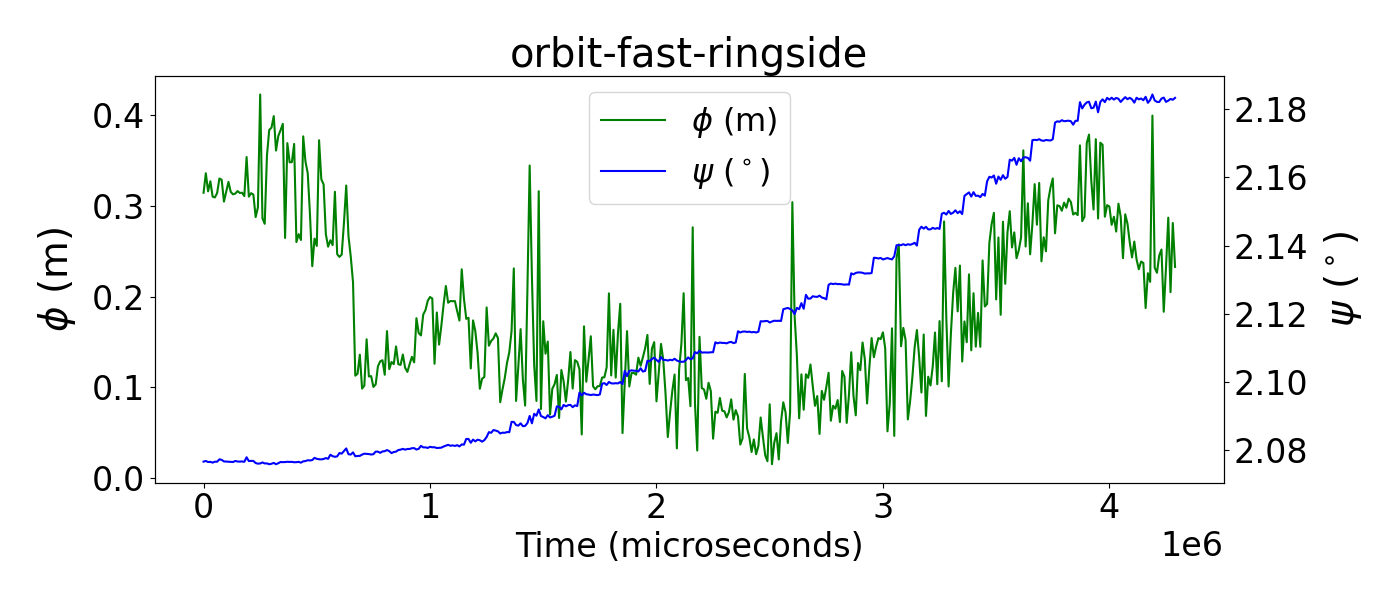}}
\caption{Relative transformation errors ($\phi_\ell$ and $\psi_\ell$) over time for several real event sequences from our dataset.}
\label{fig:quantitative_results}
\end{figure}

Fig.~\ref{fig:quantitative_results} shows the errors as a function of time of the pipeline, trained with augmentations, on 3 sequences with the best, median and worst $\Phi$. Table~\ref{tab:ablation_results} shows the overall results, which indicate excellent performance of our method since the errors were maintained at 10's of centimeters and 2-3 degrees, comparable to the top RGB methods~\cite{spec19,spec21}.

\begin{table}[ht]
\centering
\begin{NiceTabular}{llcccc}[hvlines,rules/color=[gray]{0.3}]
\Block{2-1}{} & \Block{2-1}{} & \Block{2-2}{W/O Aug.} & & \Block{2-2}{W Aug.} &\\
& & & & & \\
Trajectory & Scene & $\Phi$ (m) & $\Psi$ (${}^\circ$) & $\Phi$ (m) & $\Psi$ (${}^\circ$)\\
\Block{10-1}{approach} & fast-ambient & 0.167 & 3.011 & \textbf{0.157} & \textbf{2.090} \\
& fast-centre & 0.166 & 2.203 & \textbf{0.130} & \textbf{2.082} \\
& fast-lightbox & 0.166 & \textbf{2.923} & \textbf{0.123} & 3.156 \\
& fast-ringbelow & 0.168 & 3.054 & \textbf{0.123} & \textbf{2.449} \\
& fast-ringside & 0.153 & \textbf{3.044} & \textbf{0.131} & 3.143 \\
& slow-ambient & 0.155 & 2.166 & \textbf{0.055} & \textbf{2.075} \\
& slow-centre & 0.155 & \textbf{3.135} & \textbf{0.075} & 3.162 \\
& slow-lightbox & 0.154 & 2.974 & \textbf{0.082} & \textbf{2.086} \\
& slow-ringbelow & 0.155 & \textbf{2.800} & \textbf{0.057} & 3.140 \\
& slow-ringside & 0.155 & 2.404 & \textbf{0.116} & \textbf{2.091} \\
\Block{10-1}{orbit} & fast-ambient & 0.232 & 3.289 & \textbf{0.187} & \textbf{2.109} \\
& fast-centre & \textbf{0.189} & 2.142 & 0.200 & \textbf{2.118} \\
& fast-lightbox & 0.226 & 3.103 & \textbf{0.197} & \textbf{2.096} \\
& fast-ringbelow & 0.231 & 2.189 & \textbf{0.195} & \textbf{2.109} \\
& fast-ringside & 0.232 & 3.258 & \textbf{0.207} & \textbf{2.117} \\
& slow-ambient & 0.177 & 3.128 & \textbf{0.113} & \textbf{2.098} \\
& slow-centre & 0.222 & 3.095 & \textbf{0.119} & \textbf{2.096} \\
& slow-lightbox & 0.225 & 3.074 & \textbf{0.202} & \textbf{2.112} \\
& slow-ringbelow & 0.222 & 3.019 & \textbf{0.124} & \textbf{2.098} \\
& slow-ringside & 0.223 & 3.041 & \textbf{0.131} & \textbf{2.100} \\
\end{NiceTabular}
\caption{Ablation study showing $\Phi$ and $\Psi$ w \& w/o augmentations.}
\label{tab:ablation_results}
\end{table}

The method was challenged by event frames with low signal-to-noise ratio. We can see this at the start of \texttt{approach-slow-ambient}, when the motion was slow relative to the chosen batching duration $\tau$. Currently $\tau$ was manually chosen and fixed for each $\cE$. Automatically adjusting $\tau$ will be a fruitful research direction, \eg,~\cite{xiao2022research}.

\subsection{Ablation studies}
Our ablation results in Table.~\ref{tab:ablation_results} show that the results are generally better in the pipeline trained with augmentations.
  
\section{Conclusions}

We proposed event sensing to surmount the Sim2Real gap for satellite pose estimation. A novel event dataset with accurate ground truth was constructed to demonstrate the technique. Results show that the idea is promising. Further work will need to be done to increase the scope and breadth of the dataset, as well as raising the robustness of the method towards hyperparameter settings such as batch duration. We hope that this work drives the adoption and use of event sensors for satellite pose estimation.

\section*{Acknowledgements}

Mohsi Jawaid acknowledges support from Poppy@AIML. Ethan Elms acknowledges support from Northrop Grumman. Tat-Jun Chin is SmartSat CRC Chair of Sentient Satellites.

\pagebreak

\bibliographystyle{ieeetran}
\bibliography{references}

\end{document}